\documentclass[a4,12pt]{article}

\usepackage{amsmath,amsthm}%
\usepackage{amsfonts}%
\usepackage{amssymb}%
\usepackage{graphicx}
\usepackage{mathnotation}
\usepackage{alltt}
\usepackage{hyperref}
\usepackage{lscape}

\usepackage[draft]{fixme}
\usepackage{color}

\fxsetup{
inline,
nomargin,
theme=color
}

\newtheorem{definition}{Definition}

\title{Study: Symmetry breaking}
\author{
Technical Report\\
\
Anna Ryabokon\\
Alpen-Adria Universit\"at, Klagenfurt, Austria\\
anna.ryabokon@ifit.uni-klu.ac.at}
\date{}

\begin{document}

\maketitle

\section{Introduction}
The configuration problem consists in finding a sequence of actions required to assemble a target artifact from a set of components of predefined types. All allowed components types, their attributes and  possible relations between components are specified as configuration constraints. In addition, configuration constraints put restrictions on sets of related components required by the design of the target artifact such as a product or a service. The customization of an artifact required by the customer is formulated as customer requirements. Often these requirements can also capture customer preferences for a solution of the configuration problem. In this case the best solutions (configurations) are determined by an objective function specified in the configuration requirements. 

In their nature configuration problems are combinatorial (optimization) problems. In order to find a configuration a solver has to instantiate a number of components of a some type and each of these components can be used in a relation defined for a type. Therefore, many solutions of a configuration problem have symmetric ones which can be obtained by replacing some component of a solution by another one of the same type. These symmetric solutions decrease performance of optimization algorithms because of two reasons: a) they satisfy all requirements and cannot be pruned out from the search space; and b) existence of symmetric optimal solutions does not allow to prove the optimum in feasible time.

\section{Motivation and related work}

Symmetry breaking is fundamental topic in many AI areas based on combinatorial search. It does not matter which approach we are investigating: CSP, SAT or ASP, it is a significant tool for solving particular classes of problems. Namely, the problems which include a lot of symmetries. The pigeon-hole problem is a very good illustrating example, in which one has to place $n$ pigeons in $m$ holes such that  there is at most one pigeon in each hole. It is clear that there is no sense to distinguish between holes, because all holes are identical. Therefore, all placements of pigeons into holes belong to the same equivalence class of symmetric assignments. This is extremely important if we have $m=n-1$ holes for $n$ pigeons since one has to prove each of $(n-1)!$ branches in a search tree to verify that there is no solution. 

There are three types of symmetry breaking (SB): variable $(A,B)$, value $(A,\lnot A)$ and variable-value $(A,\lnot B)$, where $A$ and $B$ are propositional symbols and $(A,B)$ is a permutation that replaces $A$ in all clauses of a CNF with $B$ and vice versa. While the pigeon-hole problem essentially involve some sort of capacity constraint on a set of interchangeable variables, it exhibits only pure variable symmetries~\cite{satbook09}. However, breaking these symmetries improves performance a lot~\cite{Aloul2003,Aloul2006,satbook09,Drescher2011}. Although we are aware of research of SB for CSPs done e.g. by Ian P. Gent~\cite{Gent2002} and Toby Walsh~\cite{Walsh2006a}, in this study we focus us mainly on symmetry breaking for SAT problems. The reason for this is that we want to compare the approach presented in ~\cite{Friedrich2011} to the same method but extended with SB predicates.

\section{Approach}
Modern approaches to identification of symmetries in a CNF are based on a well-understood notion of group isomorphism. 
\begin{definition}[Group]
Group is a structure $\tuple{G,*}$ where $G$ is a (non-empty) set that is closed under a binary operation $*$ for which the following axioms are satisfied:
\begin{itemize}
	\item \emph{associativity}: for all $x,y,z \in G,\; (x*y)*z=x*(y*z)$
  \item \emph{identity}: there exists an element $e \in G$ such that for all $x \in G,\; x*e=x$
  \item \emph{inverse}: for each $x\in G$ there exists $x^{-1} \in G,\; x*x^{-1} = e$
\end{itemize}
\end{definition}
\noindent Note that in the literature the authors often refer $G$ to a group rather than $\tuple{G,*}$  and omit explicit definition of the operation $*$ and write $xy$ instead of $x*y$.

Let set $G=\setof{\setof{A,B}, \setof{\lnot A,B}, \setof{A, \lnot B}, \setof{\lnot A, \lnot B}}$ include sets of two propositional literals and $*: G \times G \to G$ be a binary operation defined as follows:
\begin{center}
	\begin{tabular}{l|cccc}
           	&	 $A,B$ 	&	 $\lnot A,B$ 	&	 $A, \lnot B$ 	&	 $\lnot A, \lnot B$	 \\ \hline
    $A,B$ 	&	 $A,B$ 	&	 $\lnot A,B$ 	&	 $A, \lnot B$ 	&	 $\lnot A, \lnot B$		\\
    $\lnot A,B$ 	&	 $\lnot A, B$ 	&	 $A,B$ 	&	$\lnot A, \lnot B$	&	$A, \lnot B$		\\
    $A, \lnot B$ 	&	 $A, \lnot B$ 	&	$\lnot A, \lnot B$	&	 $A,B$ 	&	$\lnot A, B$		\\
    $\lnot A, \lnot B$ 	&	 $\lnot A, \lnot B$ 	&	$A,\lnot B$	&	$\lnot A, B$	&	 $A,B$ 		\\ \hline
	\end{tabular}
\end{center}
The negation in this operation means that some element of a set should be negated. For instance, in $*(\setof{\lnot A,B},\setof{A, B})$ the first argument \setof{\lnot A,B} defines that the first element in a set should be negated and the second argument that none of the elements is negated. In this case the operation negates only $A$ and returns $\setof{\lnot A, B}$. Clearly, $\tuple{G,*}$ is a group.

\begin{definition}[Subgroup]
A group $\tuple{H,*}$ if a \emph{subgroup} of a group $\tuple{G,*}$ if $H \subseteq G$ and $H \neq \emptyset$. If $H \subset G$ then $H$ is a \emph{proper subgroup} of $G$.
\end{definition}

\begin{definition}[Group generators]
Let $H \subset G$ be a subgroup of a group $G$. The group $H$ \emph{generates} $G$ if all elements of $G$ can be obtained by (multiple)application of the group operation. Elements of $H$ are called \emph{generators} of $G$. A generator is \emph{redundant} if it can be obtained from other generators. $H$ is \emph{irredundant} if it does not contain redundant generators.
\end{definition}
An irredundant generating set of subgroup $H$ provides an extremely compact representation of $G$. Consider a group $\tuple{2\mathbb{Z},+}$ of all even integers with addition operation. In this case an irredundant set of generators $H=\setof{-2,0,2}$ provides a group $\tuple{H,+}$. 
The notion of generators provides a base for the identification of symmetries in groups. Thus, if a group $G$ contains some subgroup $G'$ which elements can be generated by its subgroup $H$ then we can consider only elements of $H$. In the context of SB for CNFs the elements of $H$ are symmetry generators. One can use $H$ to declare additional constraints eliminating symmetric solutions.

Another important notion of group theory is group isomorphism. This notion is used to relate different groups, like $G'$ and $H$ from the example given above.
\begin{definition}[Group isomorphism]
Let $\tuple{G,*}$ and $\tuple{G',*'}$ be two groups and there is an one-to-one (injective) function $\phi : G \to G'$ such that for any two elements $x,y \in G$ and corresponding elements $x',y' \in G$, i.e. $\phi(x) = x'$ and $\phi(y) = y'$:
\begin{equation}
\phi(x*y)=\phi(x) *' \phi(y) = x' *' y'
\end{equation}
then group $G$ and $G'$ are isomorphic. 
\end{definition}
\noindent That is if some property if true for the group $G$ it is also true for the group $G'$ and vice versa. Therefore, any group isomorphism maps sets of generators of a group to sets of generators of an isomorphic one.

As we mentioned above symmetric solutions of a configuration problem, and CNFs in general, are obtained by permuting either variables (propositional symbols), their values or both. 
\begin{definition}[Permutation]
A permutation $\pi$ of a set $S$ is an one-to-one and onto (bijective) function $\pi : S \to S$. Two permutations $\pi$ and $\pi'$ can be nested to form a single new permutation function by function composition $\pi''(s) = (\pi \circ \pi')(s) = \pi (\pi'(s))$, where $s \in S$.
\end{definition}

It can be easily shown that the permutation operation is bijective for any given set $S$ and therefore can be used to create a group of permutations.
% Group of permutations
\begin{definition}[Permutation Group]
Let $A$ be a non-empty set and $S_A$ be the set of all permutations of $A$. Then $S_A$ forms a group under permutation operation.
\end{definition}

Consider a simple house problem with five persons each owning five things. Given a constraint that restricts placement of things of different persons in the same cabinet, the solution of the problem includes at least five cabinets, e.g.\ $C=\setof{1,2,3,4,5}$. One of the solutions, in this case, will suggest storing all things of the first person in cabinet $1$, of the second person in the cabinet $2$ and so on. A permutation $(1,2)$ in this context means that the things of the first person will be stored in a cabinet $2$ and  things of the second person in the cabinet $1$. Identification of such permutations in a CNF formula is done through reduction to the colored graph automorphism problem. In order to define this problem let us introduce the group of a colored graph.

% Graph isomorphism / automorphism
\begin{definition}[Graph automorphism]
Given a graph $GR=(V,E)$ where $V=\setof{1,2,\dots,n}$ is a set of vertices and $E$ is a set of edges. Let $\pi(V)=\setof{V_1,V_2,\dots,V_k}$ be a partition of its vertices, i.e.\ $\bigcup_{V_i\in\pi(V)} V_i = V$ and $V_i \cap V_j = \emptyset$ for any $V_i,V_j \in \pi(V), V_i \neq V_j$. An automorphism group $Aut(GR,\pi)$ is a the set of permutations of the graph vertices of the same cell $V_i \in \pi(V)$that map edges to edges and non-edges to non-edges.
\end{definition}

% Coloring / sbass coloring schema
To simplify the presentation one can consider $\pi(V)$ as assignment of $k$ different colors to sets of graph vertices. In this case vertices of one color cannot be mapped to vertices of another one. 
% Stable coloring and generation of coloring
The coloring $\pi(V)$ is \emph{stable} if for all pairs of vertices $u,v \in V$ 
\begin{equation}
d(u,V_i) = d(v,V_i), \; \forall V_i \in \pi(V)
\end{equation} 
where $d(u,V_i)$ is a number of vertices in $V_i$ that are adjacent to $u$ in $GR$.
Given an initial coloring of a graph one can compute a set of different stable colorings. These colorings permutations of the graph vertices correspond to symmetries in the graph and, thus, form a group of permutations.

%Coloring in ASP
For the purpose of symmetry breaking in ASP Drescher et al.\ defined an initial coloring described in~\cite{Drescher2011}. A graph representation of a grounded program colored according to this definition can be used as an input to the algorithm computing stable coloring. Given a set of stable colorings it is possible to compute a set of (irredundant) generators. 
% Constraints
The latter can be used to generate a set of lexicographic constraints that introduce an order on a set of literals -- elements of the set of generators. Extension of the grounded program with these constraints leads to elimination of symmetric solutions. The meaning of these constraints can be roughly described as: literal $b$ can be in a model only if literal $a$ is.
 
\subsection{Tools}

\begin{figure}[tb]
	\centering
		\includegraphics[width=\linewidth]{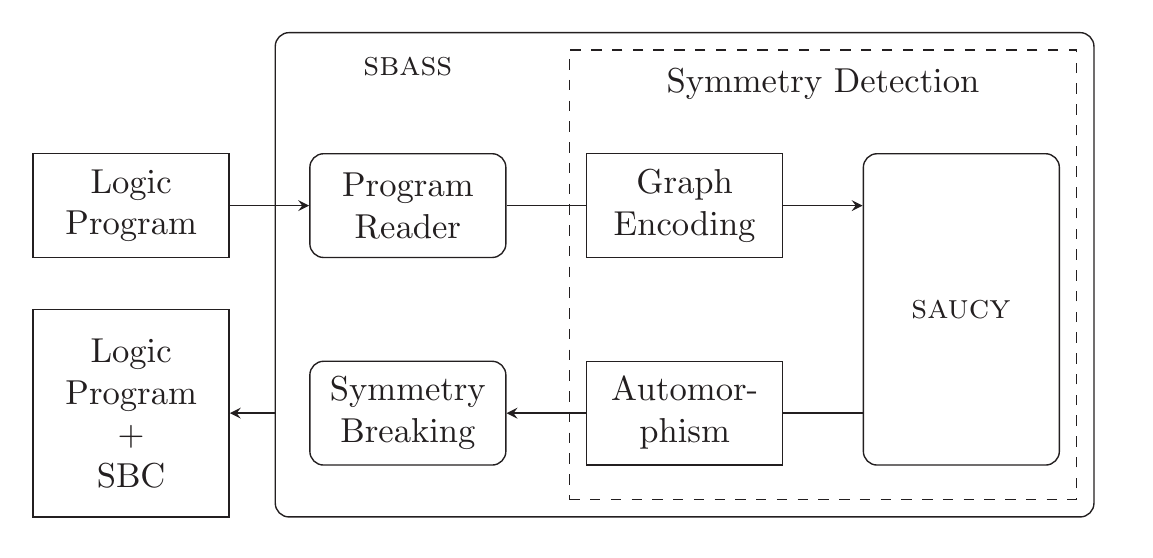}\vspace{-20pt}
	\caption{Architecture of SBASS}
	\label{fig:figArch}
  \vspace{-15pt}
\end{figure}

SBASS is a preprocessor which detects and breaks symmetries in the search space of ASP instances by adding lexicographic symmetry-breaking constraints. This tool was developed by Christian Drescher and is a part of Potsdam Answer Set Collection\footnote{\url{http://potassco.sourceforge.net/labs.html}}. SBASS takes a grounded logic program produced by a grounder GRINGO\footnote{\url{http://potassco.sourceforge.net/}} as an input. For the given grounded program the tool generates a colored graph and provides it as an input to SAUCY\footnote{\url{http://vlsicad.eecs.umich.edu/BK/SAUCY/}}~\cite{Aloul2006}. The latter is a graph automorphism identification library that returns a set of graph symmetry generators. Each symmetry generator is used to produce a chain of symmetry breaking constraints (SBC). The initial logic program is extended with SBC. The global architecture of SBASS is presented in Figure~\ref{fig:figArch}~\cite{Drescher2011}. 
SBASS allows to limit the number of computed generators\footnote{command line option \texttt{--limit=n}}, since there are exponentially many generators in the general case~\cite{satbook09}. In practice such limitation makes possible computation of symmetry breaking constraints for big grounded programs, for which identification of all symmetries would be infeasible. 

Also weight constraints and optimization is not supported by SBASS. Therefore, the weight constraint of the general encoding of the house problem provided in~\cite{Friedrich2011} was replaced by corresponding cardinality constraint and SBASS was modified to ignore optimization statement during its preprocessing step.

As a small example consider a house configuration with the following customer and configuration requirements:
\begin{alltt}\small
person(1;2).
thing(3..5).
thing(8).
personTOthing(1,3;4;5).
personTOthing(2,8).
cabinet(50;51;52;53).
1\{cabinetTOthing(X,Y):cabinet(X)\}1 :- thing(Y).
:- cabinet(X), 3{cabinetTOthing(X,Y) : thing(Y)}. % only 2 things
:- cabinetTOthing(C, T1), cabinetTOthing(C, T2), 
    personTOthing(P1,T1), personTOthing(P2, T2), P1!=P2.
\end{alltt}

Computing only one generator we set \texttt{--limit=1}. Afterwards, we obtain a group generator and a chain of SBC\footnote{It is not possible to obtain a set of SBC generated by SBASS without modification of this tool.}:
\begin{alltt}\small
(16 17) (23 24) (30 31) (37 38) (44 45) 
\end{alltt}
Translation of this output from LPARSE format~\cite{Syrjanen2000} results in the following generator set. The literals $44$ and $45$ cannot be translated since they were introduced by the grounder.
\begin{alltt}\small
(c2t(53,5)c2t(52,5)) (c2t(53,4)c2t(52,4)) 
(c2t(53,3)c2t(52,3)) (c2t(53,8)c2t(52,8)) (44 45)  
\end{alltt}
Given the generator set SBASS produces the following set of constraints: 
\begin{alltt}\small      
% thing 5 should not be placed in cabinet 53
1 1 2 1 17 16               :- not c2t(52,5), c2t(53,5).     
% all rules with cp2 in head should not be satisfied
1 1 1 0 48                  :- cp2.                          
% thing 4 in cabinet 52 should be preferred and so on...
1 48 3 1 24 23 16       cp2 :- not c2t(52,4), c2t(53,4), c2t(53,5). 
1 48 3 2 17 24 23       cp2 :- not c2t(52,5), not c2t(52,4), c2t(53,4). 
1 48 2 0 16 49          cp2 :- c2t(53,5), cp3.
1 48 2 1 17 49          cp2 :- not c2t(52,5), cp3.
1 49 3 1 31 30 23       cp3 :- not c2t(52,3), c2t(53,3), c2t(53,4).
1 49 3 2 24 31 30       cp3 :- not c2t(52,4), not c2t(52,3), c2t(53,3).
1 49 2 0 23 50          cp3 :- c2t(53,4), cp4.
1 49 2 1 24 50          cp3 :- c2t(52,4), cp4.
1 50 3 1 38 37 30       cp4 :- not c2t(52,8), c2t(53,8), c2t(53,3).
1 50 3 2 31 38 37       cp4 :- not c2t(52,3), not c2t(52,8), c2t(53,8).
1 50 2 0 30 51          cp4 :- c2t(53,3), cp5.
1 50 2 1 31 51          cp4 :- c2t(52,3), cp5.             
% do not use 44 if 45 is not used and thing 8 is in cabinet 53
1 51 3 1 45 44 37       cp5 :- not 45, 44, c2t(53,8).     
% do not use 44 if 45 is not used and thing 8 is not in cabinet 52
1 51 3 2 38 45 44       cp5 :- not c2t(52,8), not 45, 44.  


2 44 4 0 3 37 30 23 16   44 :- 3{c2t(53,8),c2t(53,3),c2t(53,4),c2t(53,5)}.
2   constraint rule
44  head
4   # literals
0   # negative literals
3   bound
37  c2t(53,8)
30  c2t(53,3)
23  c2t(53,4)
16  c2t(53,5)

1 1 1 0 44  i.e. :- 44.

2 45 4 0 3 38 31 24 17   45 :- 3{c2t(52,8),c2t(52,3),c2t(52,4),c2t(52,5)}.
2  constraint rule
45 head
4  # literals
0  # negative literals
3  bound
38 c2t(52,8)
31 c2t(52,3)
24 c2t(52,4)
17 c2t(52,5)

1 1 1 0 45 i.e. :- 45.
\end{alltt}

\subsection{Evaluation}

We evaluated\footnote{The evaluation experiments were performed using Potassco ASP collection (gringo-3.0.3, clasp-2.0.5, sbass including saucy 1.0) on a system with Intel i7-3930K CPU (3.20GHz), 32Gb of RAM and running Ubuntu 11.10.} pure application of CLASP\footnote{\url{http://potassco.sourceforge.net/}} to our general encoding and an extended by SBASS' SBC version on a set of the house reconfiguration instances where we take only creation costs for individuals into account.  

\begin{landscape}
\begin{table*}[p]
	\centering
  \scalebox{0.8}{
		\begin{tabular}{|l|c|c|c|c|c|c|}\hline
   Instance	&	Optimum	&	No SBASS	&	SBASS, default	&	SBASS, limit=5	&	SBASS, limit=10	&	SBASS, limit=20	\\	\hline    
empty\_p05t025	&	50	&	50/0:00.047	&	50/0:00.079	&	50/0:00.053	&	50/0:00.057	&	50/0:00.079	\\
empty\_p10t050	&	100	&	100/0:00.284	&	100/0:01.049	&	100/0:00.321	&	100/0:00.368	&	100/0:00.465	\\
empty\_p15t075	&	150	&	150/0:00.977	&	150/1:17.148	&	150/0:01.149	&	150/0:01.344	&	150/0:01.614	\\
empty\_p20t100	&	200	&	200/0:04.369	&	200/-	&	200/0:05.718	&	200/0:05.233	&	200/0:39.575	\\
empty\_p25t125	&	250	&	250/1:04.125	&	TO	&	250/0:58.110	&	200/1:09.729	&	250/-	\\
empty\_p30t150	&	300	&	300/-	&	TO	&	300/-	&	300/-	&	300/-	\\
empty\_p35t175	&	350	&	350/-	&	TO	&	350/-	&	350/-	&	350/-	\\
empty\_p40t200	&	400	&	400/-	&	TO	&	400/-	&	400/-	&	TO	\\ \hline 
long\_2\_p02t030c3	&	0	&	0/0:00.082	&	0/0:0.121	&	0/0:00.083	&	0/0:00.081	&	0/0:00.113	\\
long\_2\_p04t060c3	&	0	&	0/0:00.721	&	0/0:01.556	&	0/0:00.786	&	0/0:01.384	&	0/0:01.639	\\
long\_2\_p06t090c3	&	0	&	0/2:07.973	&	0/0:36.373	&	0/1:03.695	&	0/1:26.435	&	0/0:12.070	\\
long\_2\_p08t120c3	&	0	&	35/-	&	35/-	&	40/-	&	30/-	&	30/-	\\
long\_2\_p10t150c3	&	0	&	45/-	&	55/-	&	55/-	&	15/-	&	70/-	\\
long\_2\_p12t180c3	&	0	&	90/-	&	75/-	&	80/-	&	85/-	&	80/-	\\
long\_2\_p14t210c3	&	0	&	TO	&	150/-	&	TO	&	TO	&	170/-	\\
long\_2\_p16t240c3	&	0	&	TO	&	TO	&	TO	&	TO	&	TO	\\ \hline 
newroom\_p02t024c3	&	10	&	10/0:00.057	&	10/0:00.073	&	10/0:00.060	&	10/0:00.065	&	10/0:00.073	\\
newroom\_p04t048c3	&	20	&	20/0:00.398	&	20/0:00.541	&	20/0:00.483	&	20/0:00.441	&	20/0:00.446	\\
newroom\_p06t072c3	&	30	&	30/0:01.152	&	30/0:02.398	&	30/0:01.336	&	30/0:01.369	&	30/0:01.526	\\
newroom\_p08t096c3	&	40	&	40/0:02.793	&	40/0:08.380	&	40/0:03.350	&	40/0:03.485	&	40/0:03.794	\\
newroom\_p10t120c3	&	50	&	50/0:05.494	&	50/0:22.365	&	50/0:07.146	&	50/0:07.295	&	50/0:07.850	\\
newroom\_p12t144c3	&	60	&	60/0:10.073	&	60/0:53.058	&	60/0:13.488	&	60/0:13.771	&	60/0:14.749	\\
newroom\_p14t168c3	&	70	&	70/0:16.827	&	70/2:00.135	&	70/0:23.840	&	70/0:24.258	&	70/0:25.768	\\
newroom\_p16t192c3	&	80	&	80/0:24.816	&	80/3:52.280	&	80/0:38.284	&	80/0:38.753	&	80/0:41.115	\\ \hline 
swap\_r02t035	&	0	&	0/0:00.038	&	0/0:00.064	&	0/0:00.046	&	0/0:00.048	&	0/0:00.048	\\
swap\_r04t070	&	0	&	0/0:00.124	&	0/0:00.250	&	0/0:00.152	&	0/0:00.155	&	0/0:00.153	\\
swap\_r06t105	&	0	&	0/0:00.279	&	0/0:00.964	&	0/0:00.343	&	0/0:00.344	&	0/0:00.351	\\
swap\_r08t140	&	0	&	0/0:00.594	&	0/0:02.855	&	0/0:00.724	&	0/0:00.734	&	0/0:00.743	\\
swap\_r10t175	&	0	&	0/0:01.112	&	0/0:06.655	&	0/0:01.328	&	0/0:01.278	&	0/0:01.324	\\
swap\_r12t210	&	0	&	0/0:01.853	&	0/0:16.417	&	0/0:02.171	&	0/0:02.151	&	0/0:02.224	\\
swap\_r14t245	&	0	&	0/0:02.721	&	0/0:28.879	&	0/0:03.406	&	0/0:03.353	&	0/0:03.496	\\
swap\_r16t280	&	0	&	0/0:04.407	&	0/0:48.896	&	0/0:05.115	&	0/0:05.299	&	0/0:05.196	\\   \hline
		\end{tabular}   
     }    
	\caption{Evaluation results for the house reconfiguration problem. $TO$ indicates a timeout within 600 seconds. $50/0:00.047$ means that an optimal solution with costs equal 50 was found in 47 milliseconds, whereas $200/-$ reports that only a suboptimal solution with costs equal 200 was returned by a solver.}
	\label{tab:eval}
  \vspace{-10pt}
\end{table*}
\end{landscape}

We tested application of SBASS with default settings when all generators have to be computed and we limited search of generators by a constant. We set this constant to 5, 10 and 20 during the evaluation to see how it influences on performance . Overall evaluation results are presented in Table~\ref{tab:eval}.

\section{Conclusions}

The results for the pigeon-hole problem presented in~\cite{Drescher2011} are very impressive. This motivated us to try symmetry breaking tool suitable for ASP suggested there. Unfortunately, it turned out that not all of the house reconfiguration problem instances can be solved in a given time frame although we limit a number of generators. Moreover, in only 2 cases (empty\_p25t125, long\_2\_p06t090c3) runtime was improved and in 3 cases (long\_2\_p08t120c3, long\_2\_p10t150c3, long\_2\_p12t180c3) CLASP found the better suboptimal solutions by application of SBASS. Together, there are only 5 cases from 32 were actually runtime or quality of a solution was better by adding of SBC. The reason for this could be that the size of SBC is too large to be effectively handled by a SAT solver~\cite{satbook09} on the one hand. On the other hand, additional constraints, like we are not allowed to store things of different persons in the same cabinet, might cause the difficulties. 

Some other modern packages for detecting and breaking symmetries of CNF formulas are available. NAUTY\footnote{\url{http://potassco.sourceforge.net/labs.html}} described in~\cite{Aloul2003} is another approach to compute automorphism groups of graphs. Experiments showed that it is not efficient enough for large sparse but for dense graphs~\cite{satbook09}. The instances of the house (re)configuration problem are sparse graphs and choice of NAUTY would be not justified. Junttila et Al~\cite{Junttila2007} introduced BLISS\footnote{\url{http://www.tcs.hut.fi/Software/bliss/index.html}} which is an enhancement of NAUTY and SAUCY. The authors showed experimentally that their approach outperforms the previous tools. However, these were not investigated in this study.

%references: ~\cite{Aloul2006,Junttila2007,satbook09,Aloul2003,Katebi2010,Drescher2011,Ramani2006,Friedrich2011}, 

\bibliographystyle{plain}
\bibliography{sym_break}
\end{document}